# Feature-based Recognition Framework for Super-resolution Images

JingHu[1] Meiqi Zhang[1]* Rui Zhang[1]

1 School of Artificial
Intelligence and Automation.HUST

## Abstract

In practical application, the performance of recognition network usually decreases when being applied on super-resolution images. In this paper, we propose a feature-based recognition network combined with GAN (FGAN). Our network improves the recognition accuracy by extracting more features that benefit recognition from SR images. In the experiment, we build three datasets using three different super-resolution algorithm, and our network increases the recognition accuracy by more than 6% comparing with ReaNet50 and DenseNet121.

## 1 INTRODUCTION

In the past years, deep neural networks (DNNs) have been widely applied in the field of image recognition [1-4]. Most of these image recognition models are trained and evaluated on high-resolution (HR) images. However, in practical application, the input of recognition is usually low-resolution (LR) images, like video surveillances and medical images. In this case, image super-resolution (SR) is a popular solution [5-7].

Super-resolution means finding a mapping from the LR image to its HR version [8]. Recently, many SR models based on deep learning have been introduced. In [9], Dong C proposes SR-CNN method where convolutional neural network (CNN) is applied to the field of image super-resolution for the first time. In [10], SRRESNET method is established which combines global residual learning and local residual learning to learn more global feature information. In [11], Wang X introduces generative adversarial network (GAN) into the field of super-resolution reconstruction and proposes SRGAN method based on GAN. These SR methods use different network architectures to improve the SR performance by improving the Peak Signal-to-Noise Ratio (PSNR) value.

Even though these super resolution reconstructions can be visually appealing, most SR methods are not designed for image recognition. These PSNR-oriented SR methods tend to produce over-smoothed output and lose key features. Meanwhile, the recognition networks commonly used, such as ResNet50, VGG-16 and DenseNet121, classify images according to their features [2, 12-13]. Therefore, the recognition performances of SR images decrease significantly compared with that of HR images. To solve this problem, the existing methods are conducting super-resolution in feature domain, such as in [5,14,16], instead of making improvements in the process of recognition.

In this paper, we propose a feature-based recognition network combined with GAN (FGAN). Considering that the decline of recognition accuracy of SR image is due to the lack of features compared with HR images, we construct a network structure to extract more features that benefit recognition from SR images.

The main advantages of our framework can be summarized as follows:

1. The main idea of our framework is to optimize the feature extraction network of the image recognition method, so that the method proposed does not need to limit the super-resolution algorithm used.
2. Our framework is easy to accomplish, since it does not change the structure of recognition network. And the training process of our framework is efficient because of the usage of pre-trained recognition network.
3. The experiment results show that the recognition effect of proposed framework on SR images is extremely close to that on HR images.

## 2 RELATED WORK

**2.1 Recognition of Super Resolution Images**

In the field of face recognition, gesture recognition in surveillance video and so on, recognition of super resolution images is commonly applied due to the low resolution of the original images. In [5], Gunturk proposes to transfer the super-resolution reconstruction from pixel domain to a lower dimensional face space. In [14], K. Nguyen proposes a feature-domain super-resolution framework for Gabor-based face and iris recognition. This framework improves the recognition performance of SR images by conducting super-resolution in the non-linear Gabor feature domain. Experiments have shown the proposed approach in [14] demonstrates superior performance for both face and iris biometrics. In [16], Noor developed a machine vision tasks-friendly super-resolution technique which enhances the gradient images and associated features from the low-resolution images that benefit the high level machine vision tasks. Simulation results demonstrate the performance gains in both gradient image quality as well as key points repeatability.

All the above methods try to improve the recognition performance of SR images through reconstructing SR images in feature domain, and they are making some progress.

However, some features have been lost in LR images during the degradation process, and super-resolution algorithm cannot reconstruct these features, even if reconstructing degraded images in feature domain. We suppose it is more effective to use features that still exist in the SR images and are similar to the corresponding HR images. Therefore, we develop a recognition framework for SR images which can extract similar features from SR images as HR images.

**2.2 Generative Adversarial Network**

We borrowed the idea of Generative adversarial network (GAN) in our framework. GAN is first introduced by Goodfellow in [17], which proposes a minimax two-player game between a generative model G and a discriminative model D. Model G captures the data distribution, and model D estimates the probability that a sample came from the training data rather than G. D and G play the two-players minimax game with value function $V(G, D)$:

$$\min_G \max_D V(D, G) = \mathbb{E}_{x \sim p_{data}(x)}[\log D(x)] + \mathbb{E}_{z \sim p_z(z)}\left[\log\left(1 - D(G(z))\right)\right]. \quad (1)$$

where $D(x)$ represents the probability that the model D discriminates data x as a real sample. $p_{data}$ and $p_z$ are the probability distributions of data x and random noise z, respectively.

However, the vanilla GAN is difficult to train since the generator accepts random noise as input and tries to model the probability distributions of real data. The conditional GAN, introduced in [18], manages to mitigate this problem. The conditional GAN stabilizes the training process by accessing extra information from conditional input. The cost function of conditional GAN is given by:

$$\min_G \max_D V(D,G) = \mathbb{E}_{x\sim p_{data}(x)}[\log D(x|y)] +$$
$$\mathbb{E}_{z\sim p_z(z)}\left[\log\left(1 - D(G(z|y))\right)\right]. \quad (2)$$

where y represents the conditional input. Conditional GAN has been widely applied to various image-to-image translation tasks [19-21] and achieves satisfactory performance.

Our framework is inspired by the structure of conditional GAN, but we change the discriminator structure to be the same as a classifier part of a CNN based recognition network, because the input of our discriminator is features extracted from images and our discriminator needs to estimate input features based on the classification effect.

## 3 PROPOSED METHOD
### 3.1 Framework Structure

Our framework structure is shown in Figure 2. We define a CNN based image recognition model $M$ with $d$ layers. The model $M$ can be divided into feature extraction module $M_{fe}$ and classification module $M_c$. This division strategy is applicable for most of the current mainstream recognition networks [12-13,22].

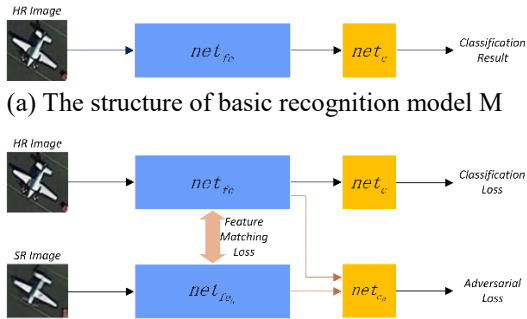

(a) The structure of basic recognition model M

(b) The structure of our recognition model FGAN

Figure 1 The structure of recognition model

The structure of our proposed framework is shown in Figure 2. This framework consists of four parts: generator $net_{fe_G}$, feature extraction network $net_{fe}$, classifier $net_c$ and discriminator $net_{c_D}$.

### 3.2 Loss Function

The proposed framework is trained on SR images $\{y_i\}$ by a combined loss $L_{comb}$ as given by:
$$L_{comb} = \lambda_{cls}L_{cls} + \lambda_{fe}L_{fe} + \lambda_{adv}L_{adv} \quad (3)$$
where $L_{cls}$ is the classification loss, $L_{fe}$ is the feature matching loss, $L_{adv}$ is the adversarial loss.

## 4 EXPERIMENTS
### 4.1 Dataset Setup

We evaluated our model on DOTA-v1.5 dataset [25]. It contains 0.4 million annotated object instances within 16 categories, which include plane, ship, storage tank, baseball diamond, tennis court, basketball court, ground track field, harbor, bridge, small vehicle, large vehicle, helicopter, roundabout, soccer ball field, swimming pool and container crane. We select the 7 categories that contain more than 500 images to be used as HR images in the training and testing process. And all the images used are under 150×150 in size.

We get the LR images by down-sampling HR images using MATLAB bicubic kernel function with a scaling factor of ×4. Then we generate the corresponding SR images using three different super-resolution algorithm: residual channel attention networks (RCAN) [23], enhanced super-resolution generative adversarial networks (ESRGAN) [11] and enhanced deep residual networks (EDSR) [24]. We mark these three SR datasets as RCAN-SR, ESRGAN-SR and EDSR-SR, respectively. The super-resolution results are shown in Figure 2.

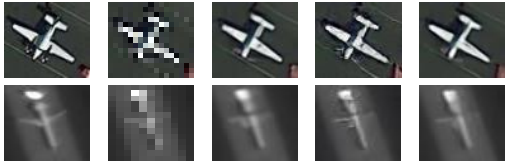

(a) HR image  (b) LR image  (c) RCAN-SR image  (d) ESRGAN-SR image  (e) EDSR-SR image

Figure 2 Examples of the HR and SR datasets

2.2 Image recognition on the subset of DOTA-v1.5

## 4.2 Image recognition on the subset of DOTA-v1.5

We adopt two mainstream image recognition models ResNet50 [2] and DenseNet121 [13] as base model $M$. The fully connected layer in $M$ is regarded as the classification module $M_c$ and the rest of $M$ is the feature extraction module $M_{fe}$. We adopt a pre-trained model to accelerate the convergence, and use the same data augmentation strategy during the training process. in an alternative way. We use stochastic gradient descent as optimizer and set the momentum to 0.99. All learning rates of our framework use an exponential decay of 0.95. We set batch size as 16 and the maximum number of epochs as 200. We use early stopping when the validation loss has not been updated for five consecutive epochs. We compare the following approaches:

HR: M is trained on $\{\{x_i\}, \{l_i\}\}$ and tested on $\{\{y_i\}, \{l_i\}\}$

SR: M is trained on $\{\{y_i\}, \{l_i\}\}$ and tested on $\{\{y_i\}, \{l_i\}\}$

HR-SR: M is trained on $\{\{x_i\}, \{l_i\}\}$ and $\{\{y_i\}, \{l_i\}\}$ at the same time and tested on $\{\{y_i\}, \{l_i\}\}$

FGAN: M is trained using our proposed framework and tested on $\{\{y_i\}, \{l_i\}\}$

Table 1 shows the results of these recognition networks on three different SR datasets using ResNet50 as M. Table 2 shows the results of these recognition networks on three different SR datasets using DenseNet121 as M. For comparison, we added the recognition result of the model M trained on $\{\{x_i\}, \{l_i\}\}$ and tested on $\{\{x_i\}, \{l_i\}\}$, marked as **HRori**.

It is noteworthy that for all three SR datasets, the recognition accuracy of FGAN is better than any other approach, when the base model M is ResNet50 or DenseNet121. And the classification effect of FGAN on SR images is extremely close to that of HRori on HR images.

Table 1 Recognition results of model based on ResNet50

| Dataset | Approach | Accuracy | F1 Scores |
|---|---|---|---|
| HR | HRori | 90.61% | 0.9121 |
| RCAN-SR | HR | 84.75% | 0.8561 |
|  | SR | 82.85% | 0.8563 |
|  | HR-SR | 85.16% | 0.8724 |
|  | FGAN | **91.30%** | **0.9255** |
| ESRGAN-SR | HR | 82.91% | 0.8454 |
|  | SR | 84.08% | 0.8576 |
|  | HR-SR | 86.28% | 0.8763 |
|  | FGAN | **91.35%** | **0.9227** |
| EDSR-SR | HR | 84.93% | 0.8538 |
|  | SR | 83.99% | 0.8679 |
|  | HR-SR | 85.96% | 0.8718 |
|  | FGAN | **91.97%** | **0.9250** |

Table 2 Recognition results of model based on DenseNet121

| Dataset | Approach | Accuracy | F1 Scores |
|---|---|---|---|
| HR | HRori | 92.18% | 0.9254 |
| RCAN-SR | HR | 85.32% | 0.8620 |
|  | SR | 86.50% | 0.8832 |
|  | HR-SR | 87.62% | 0.8892 |
|  | FGAN | **93.26%** | **0.9375** |
| ESRGAN-SR | HR | 87.04% | 0.8834 |
|  | SR | 84.73% | 0.8660 |
|  | HR-SR | 87.27% | 0.8832 |
|  | FGAN | **93.12%** | **0.9380** |
| EDSR-SR | HR | 84.86% | 0.8538 |
|  | SR | 86.91% | 0.8799 |
|  | HR-SR | 87.82% | 0.8861 |
|  | FGAN | **92.13%** | **0.9307** |

## 5 CONCLUSION AND PROSPECT

In practical application, the recognition effect of state-of-the-art networks trained on high quality images decays when being tested on super-resolution images. In this work, we find that the classification accuracy can be improved while we optimize the feature extraction module. Therefore, we propose a novel framework that can optimize the feature extraction network to get features more effective to classify. Our framework uses the structure of conditional GAN, but we change the structure of discriminative so that it can be more suitable to estimates features instead of original data such as images and noises. We thoroughly evaluate our proposed framework on datasets generated by various super-resolution algorithm and recognition networks, which confirms the effectiveness and generalization of our method.

Furthermore, considering that the existing recognition for SR images is optimized in the super-resolution process, we can try to combine the FGAN proposed in this paper with those existing recognition methods. This probably can increase the recognition effect on the basis of their super-resolution recognition effect.

## References

[1] Brock, A., De, S., Smith, S. L., & Simonyan, K. (2021). High-performance large-scale image recognition without normalization. arXiv preprint arXiv:2102.06171.

[2] Simonyan, K., & Zisserman, A. (2014). Very deep convolutional networks for large-scale image recognition. arXiv preprint arXiv:1409.1556.

[3] Cai, Q., Pan, Y., Yao, T., Yan, C., & Mei, T. (2018). Memory matching networks for one-shot image recognition. In Proceedings of the IEEE conference on computer vision and pattern recognition (pp. 4080-4088).

[4] Li, Y., Luo, C., Yang, H., & Wu, T. (2019, July). Convolutional Neural Networks for Scene Image Recognition. In International Conference on Artificial Intelligence and Security (pp. 463-474). Springer, Cham.

[5] Gunturk, B. K., Batur, A. U., Altunbasak, Y., Hayes, M. H., & Mersereau, R. M. (2003). Eigenface-domain super-resolution for face recognition. IEEE transactions on image processing, 12(5), 597-606.

[6] Huang, H., & He, H. (2010). Super-resolution method for face recognition using nonlinear mappings on coherent features. IEEE Transactions on Neural Networks, 22(1), 121-130.

[7] Rasti, P., Uiboupin, T., Escalera, S., & Anbarjafari, G. (2016, July). Convolutional neural network super resolution for face recognition in surveillance monitoring. In International conference on articulated motion and deformable objects (pp. 175-184). Springer, Cham.

[8] Siu, W. C., & Hung, K. W. (2012, December). Review of image interpolation and super-resolution. In Proceedings of The 2012 Asia Pacific Signal and Information Processing Association Annual Summit and Conference (pp. 1-10). IEEE.

[9] Dong, C., Loy, C. C., He, K., & Tang, X. (2015). Image super-resolution using deep convolutional networks. IEEE transactions on pattern analysis and machine intelligence, 38(2), 295-307.

[10] Ledig, C., Theis, L., Huszár, F., Caballero, J., Cunningham, A., Acosta, A., ... & Shi, W. (2017). Photo-realistic single image super-resolution using a generative adversarial network. In Proceedings of the IEEE conference on computer vision and pattern recognition (pp. 4681-4690).

[11] Wang, X., Yu, K., Wu, S., Gu, J., Liu, Y., Dong, C., ... & Change Loy, C. (2018). Esrgan: Enhanced super-resolution generative adversarial networks. In Proceedings of the European conference on computer vision (ECCV) workshops (pp. 0-0).

[12] He, K., Zhang, X., Ren, S., & Sun, J. (2016). Deep residual learning for image recognition. In Proceedings of the IEEE conference on computer vision and pattern recognition (pp. 770-778).

[13] Huang, G., Liu, Z., Van Der Maaten, L., & Weinberger, K. Q. (2017). Densely connected convolutional networks. In Proceedings of the IEEE conference on computer vision and pattern recognition (pp. 4700-4708).

[14] Nguyen, K., Sridharan, S., Denman, S., & Fookes, C. (2012, June). Feature-domain super-resolution framework for Gabor-based face and iris recognition. In 2012 IEEE Conference on Computer Vision and Pattern Recognition (pp. 2642-2649). IEEE.

[15] Sun, Z., Ozay, M., Zhang, Y., Liu, X., & Okatani, T. (2018). Feature quantization for defending against distortion of images. In Proceedings of the IEEE Conference on Computer Vision and Pattern Recognition (pp. 7957-7966).

[16] Noor, D. F., Li, Y., Li, Z., Bhattacharyya, S., & York, G. (2019, May). Gradient image super-resolution for low-resolution image recognition. In ICASSP 2019-2019 IEEE International Conference on Acoustics, Speech and Signal Processing (ICASSP) (pp. 2332-2336). IEEE.

[17] Goodfellow, I., Pouget-Abadie, J., Mirza, M., Xu, B., Warde-Farley, D., Ozair, S., ... & Bengio, Y. (2014). Generative adversarial nets. Advances in

neural information processing systems, 27.

[18] Mirza, M., & Osindero, S. (2014). Conditional generative adversarial nets. arXiv preprint arXiv:1411.1784.

[19] Antipov, G., Baccouche, M., & Dugelay, J. L. (2017, September). Face aging with conditional generative adversarial networks. In 2017 IEEE international conference on image processing (ICIP) (pp. 2089-2093). IEEE.

[20] Chen, C., Mu, S., Xiao, W., Ye, Z., Wu, L., & Ju, Q. (2019, July). Improving image captioning with conditional generative adversarial nets. In Proceedings of the AAAI Conference on Artificial Intelligence (Vol. 33, No. 01, pp. 8142-8150).

[21] Isola, P., Zhu, J. Y., Zhou, T., & Efros, A. A. (2017). Image-to-image translation with conditional adversarial networks. In Proceedings of the IEEE conference on computer vision and pattern recognition (pp. 1125-1134).

[22] Krizhevsky, A., Sutskever, I., & Hinton, G. E. (2012). Imagenet classification with deep convolutional neural networks. Advances in neural information processing systems, 25, 1097-1105.

[23] Zhang, Y., Li, K., Li, K., Wang, L., Zhong, B., & Fu, Y. (2018). Image super-resolution using very deep residual channel attention networks. In Proceedings of the European conference on computer vision (ECCV) (pp. 286-301).

[24] Lim, B., Son, S., Kim, H., Nah, S., & Mu Lee, K. (2017). Enhanced deep residual networks for single image super-resolution. In Proceedings of the IEEE conference on computer vision and pattern recognition workshops (pp. 136-144).

[25] https://captain-whu.github.io/DOTA/dataset.html